\documentclass[11pt]{article}

\PassOptionsToPackage{numbers, sort&compress}{natbib}

\usepackage{comment}
\usepackage{tabularx}

\usepackage[table]{xcolor}

\newcommand{\verylightgreen}{blue!5} 

\usepackage[notheorems]{rr-math}

\usepackage{paralist}

\providecommand{\y}{\ensuremath{w}} 

\providecommand{\nL}{\ensuremath{L}}  

\newcommand{\eg}{\emph{e.g.}}

\usepackage[final]{nips_2016}

\usepackage{booktabs}       
\usepackage{nicefrac}       
\usepackage{microtype}      

\title{Inductive Representation Learning \\in Large Attributed Graphs}

\author{
Nesreen K. Ahmed \\
Intel Labs\\
\texttt{nesreen.k.ahmed@intel.com} \\
\And
Ryan A. Rossi\\
PARC\\
\texttt{rrossi@parc.com}\\
\And
Rong Zhou\\
Google\\
\texttt{rongzhou@google.com}\\
\And
John Boaz Lee\\
WPI\\
\texttt{jtlee@wpi.edu}\\
\And
Xiangnan Kong\\
WPI\\
\texttt{xkong@wpi.edu}\\
\And
Theodore L. Willke\\
Intel Labs \\
\texttt{ted.willke@intel.com}\\
\And
Hoda Eldardiry\\
PARC\\
\texttt{heldardiry@parc.com}
}

\begin{document}

\maketitle

Graphs (networks) are ubiquitous and allow us to model entities (nodes) and the dependencies (edges) between them.
Graph data is often observed directly in the natural world (\eg, biological or social networks)~\cite{viswanath2009evolution} or constructed from non-relational data by deriving a metric space between entities and retaining only the most significant edges~\cite{zhu2003semi,henaff2015deep}.
Learning a useful feature representation from graph data lies at the heart and success of many machine learning tasks such as classification, anomaly detection, link prediction, 
among many others.
Many existing techniques use \emph{random walks} as a basis for learning features or estimating the parameters of a graph model for a downstream prediction task.
Examples include recent node embedding methods such as DeepWalk~\cite{deepwalk}, node2vec~\cite{node2vec}, as well as graph-based deep learning algorithms.
However, the simple random walk used by these methods is fundamentally tied to the \emph{identity} of the node.
This has three main disadvantages.
First, these approaches are inherently transductive and do not generalize to unseen nodes and other graphs~\cite{deepGL}.
Second, they are not space-efficient as a feature vector is learned for each node which is impractical for large graphs.
Third, most of these approaches lack support for \emph{attributed graphs}.

To make these methods more generally applicable, 
we propose a framework for \emph{inductive network representation learning} based on the notion of attributed random walk that is not tied to node identity and is instead based on learning a function $\phi : \vx \rightarrow \y$ that maps a node attribute vector to a type.
This framework serves as a basis for generalizing existing methods such as DeepWalk, node2vec, LINE, and many other existing (and future) methods that leverage traditional random walks.
Given an (un)directed graph $G=(V,E)$, the framework consists of two general steps:
\begin{enumerate}
\item[$\mathbf{1}$.] \textbf{Function Mapping Nodes to Types}:
The first step is to learn a function $\phi$ that maps nodes to types based on a $N \times K$ matrix $\mX$ of attributes.
Note $\mX$ may be given as input and/or computed based on the structure of the graph.

\item[$\mathbf{2}$.] \textbf{Attributed Random Walks}: 
The second step uses the types derived by the function $\phi$ for generating  \emph{attributed random walks}.
An \emph{attributed walk} $S$ of length $\nL$ is defined as a sequence of adjacent node types 
$\phi(\vx_{i_{1}}), \phi(\vx_{i_{2}}),\ldots, \phi(\vx_{i_{\nL+1}})$
associated with a sequence of indices $i_{1}, i_{2}, \ldots, i_{\nL+1}$ such that $(v_{i_{t}}, v_{i_{t+1}}) \in E$ for all $1 \leq t \leq \nL$.
\end{enumerate}
The set of attributed random walks can then be given as input into the Skip-Gram model (or other representation learning methods) to learn embeddings for the node types (as opposed to the nodes themselves).
Our proposed framework has the following key properties:

{
\begin{compactitem}
\setlength{\parskip}{1mm}
\item \textbf{Space-efficient}: It requires on average $853$x less space than existing methods.

\item \textbf{Accurate}: It is accurate with an average improvement of $16.1\%$ across a variety of graphs from several domains.

\item \textbf{Inductive}: 
It is an inductive network representation learning approach that is able to learn embeddings for 
new nodes and graphs.

\item \textbf{Attributed}: 
It naturally supports graphs with attributes (if available) and serves as a foundation for generalizing existing methods for use on attributed graphs.

\vspace{1.5mm}
\end{compactitem}
}

\newcolumntype{H}{>{\setbox0=\hbox\bgroup}c<{\egroup}@{}}
\newcommand{\dataName}[1]{\ensuremath{\fontsize{6}{7.5}\selectfont \mathsf{#1}}} 
\newcommand{\ra}[1]{\renewcommand{\arraystretch}{#1}} 

\begin{table}[h!]
\centering
\setlength{\tabcolsep}{6.5pt}
\ra{1.1}
\scriptsize
\caption{
AUC scores for Link Prediction. See text for discussion. 
}
\vspace{1mm}
\label{table:link-pred-mean-OP}
\small
\fontsize{8.0}{9.0}\selectfont
\begin{tabularx}{0.8\linewidth}{r H cccc H} 
\toprule
\textsc{Graph}  &&
\textbf{Our Method} & \textbf{Node2Vec}~\cite{node2vec} & \textbf{DeepWalk}~\cite{deepwalk} & \textbf{LINE}~\cite{line} &
\\
\midrule

\dataName{biogrid\text{--}human}  && 
\cellcolor{\verylightgreen} \text{$\mathbf{0.877}$}  &  
0.869  &  
0.864  &  
0.744  &  
\\

\dataName{bn\text{--}cat}  && 
\cellcolor{\verylightgreen} \text{$\mathbf{0.710}$}  &  
0.627  &  
0.627  &  
0.672  &  
\\

\dataName{bn\text{--}rat\text{--}brain}  && 
\cellcolor{\verylightgreen} \text{$\mathbf{0.748}$}  &  
0.716  &  
0.716  &  
0.691  &  
\\

\dataName{bn\text{--}rat\text{--}cerebral}  && 
\cellcolor{\verylightgreen} \text{$\mathbf{0.867}$}  &  
0.813  &  
0.811  &  
0.709  &  
\\

\dataName{ca\text{--}CSphd}  && 
\cellcolor{\verylightgreen} \text{$\mathbf{0.838}$}  &  
0.768  &  
0.735  &  
0.620  &  
\\

\dataName{eco\text{--}everglades}  && 
\cellcolor{\verylightgreen} \text{$\mathbf{0.762}$}  &  
0.739  &  
0.739  &  
0.704  &  
\\

\dataName{eco\text{--}fweb\text{--}baydry}  && 
\cellcolor{\verylightgreen} \text{$\mathbf{0.681}$}  &  
0.655  &  
0.627  &  
0.660  &  
\\

\dataName{ia\text{--}radoslaw\text{--}email}  && 
\cellcolor{\verylightgreen} \text{$\mathbf{0.867}$}  &  
0.756  &  
0.745  &  
0.769  &  
\\

\dataName{inf\text{--}USAir97}  && 
\cellcolor{\verylightgreen} \text{$\mathbf{0.884}$}  &  
0.881  &  
0.834  &  
0.843  &  
\\

\dataName{soc\text{--}anybeat}  && 
\cellcolor{\verylightgreen} \text{$\mathbf{0.961}$}  &  
0.854  &  
0.848  &  
0.850  &  
\\

\dataName{soc\text{--}dolphins}  && 
\cellcolor{\verylightgreen} \text{$\mathbf{0.656}$}  &  
0.580  &  
0.498  &  
0.551  &  
\\

\dataName{fb\text{--}Yale4}  && 
\cellcolor{\verylightgreen} \text{$\mathbf{0.793}$}  &  
0.742  &  
0.728  &  
0.763  &  
\\

\dataName{fb\text{--}nips\text{--}ego}  && 
\cellcolor{\verylightgreen} \text{$\mathbf{0.998}$}  &  
0.997  &  
0.996  &  
0.743  &  
\\

\dataName{web\text{--}EPA}  && 
\cellcolor{\verylightgreen} \text{$\mathbf{0.926}$}  &  
0.804  &  
0.738  &  
0.768  &  
\\

\midrule

\end{tabularx}
\end{table}

To evaluate the effectiveness of the proposed framework we use it to generalize node2vec.
Additional details are discussed in~\cite{ahmed17Gen-Deep-Graph-Learning}.
Experimental results are provided in Table~\ref{table:link-pred-mean-OP}.
The results in Table~\ref{table:link-pred-mean-OP} use the \textsc{mean op} $(\vz_i + \vz_j)\big/2$ to construct edge features from the learned node embedding vectors $\vz_i$ and $\vz_j$ of node $i$ and $j$. Similar results were observed using other operators.
In all cases, the generalized approach outperforms all other methods across a wide variety of networks from biology to information and social networks.
The experimental results in Table~\ref{table:link-pred-mean-OP} demonstrate the effectiveness of the proposed framework for generalizing existing methods making them more powerful and practical for attributed and heterogeneous/typed networks~\cite{kong2012meta,sun2013mining,sun2012mining} as well as for inductive learning tasks~\cite{zager2008graph,singh2007-matching-topology,neville:kdd03,roles2015-tkde,holme2012temporal,deepGL,oyama2011cross}.
Notably, the generalization enables these methods to leverage any available intrinsic or self-attributes as well as any arbitrary set of structural features derived from the graph (\eg, higher-order subgraph features such as graphlets~\cite{pgd}) or even relational features that leverage the graph structure as well as the initial intrinsic/self-attributes~\cite{rossi12jair}.
Furthermore, the learned embeddings from our approach naturally generalize as they are no longer tied to identity and instead represent general functions that can easily be extracted on another arbitrary graph. Therefore, the approach naturally supports a wide range of \emph{inductive network representation learning}~\cite{deepGL} tasks
such as cross-temporal link prediction~\cite{oyama2011cross}, classification in dynamic networks~\cite{parkdeep}, across-network prediction and modeling~\cite{neville:kdd03}, role discovery~\cite{roles2015-tkde}, graph matching~\cite{singh2007-matching-topology} \& similarity~\cite{zager2008graph}, 
among others~\cite{akoglu2015graph,lee17-Deep-Graph-Attention,ahmed17Gen-Deep-Graph-Learning}.
Finally, the generalized approach using the proposed framework is guaranteed to perform at least as well as the original method since it is recovered as a special case of the framework.
Other advantages of the framework are discussed in the full version of this paper~\cite{ahmed17Gen-Deep-Graph-Learning}.

\noindent
\textbf{Conclusion}:
This work proposed a flexible framework for generalizing an important class of embedding and representation learning methods for graphs that leverage random walks.
The framework serves as a basis for generalizing existing methods for use with attributed graphs, unseen nodes, inductive learning/graph-based transfer learning tasks, while also allowing significantly larger graphs due to the inherent space-efficiency of the approach.
Finally, the framework was shown to have the following desired properties: space-efficient, accurate, inductive, and able to support graphs with attributes (and more generally heterogeneous/typed networks).

\setlength{\bibsep}{4pt}
{\small
\fontsize{10pt}{11pt}\selectfont
\bibliography{paper}}

\begin{thebibliography}{22}
\providecommand{\natexlab}[1]{#1}
\providecommand{\url}[1]{\texttt{#1}}
\expandafter\ifx\csname urlstyle\endcsname\relax
  \providecommand{\doi}[1]{doi: #1}\else
  \providecommand{\doi}{doi: \begingroup \urlstyle{rm}\Url}\fi

\bibitem[Ahmed et~al.(2015)Ahmed, Neville, Rossi, and Duffield]{pgd}
N.~K. Ahmed, J.~Neville, R.~A. Rossi, and N.~Duffield.
\newblock Efficient graphlet counting for large networks.
\newblock In \emph{ICDM}, page~10, 2015.

\bibitem[Ahmed et~al.(2017)Ahmed, Rossi, Zhou, Lee, Kong, Willke, and
  Eldardiry]{ahmed17Gen-Deep-Graph-Learning}
N.~K. Ahmed, R.~A. Rossi, R.~Zhou, J.~B. Lee, X.~Kong, T.~L. Willke, and
  H.~Eldardiry.
\newblock A framework for generalizing graph-based representation learning
  methods.
\newblock In \emph{arXiv:1709.04596}, 2017.

\bibitem[Akoglu et~al.(2015)Akoglu, Tong, and Koutra]{akoglu2015graph}
L.~Akoglu, H.~Tong, and D.~Koutra.
\newblock Graph based anomaly detection and description: a survey.
\newblock \emph{DMKD}, 29\penalty0 (3):\penalty0 626--688, 2015.

\bibitem[Grover and Leskovec(2016)]{node2vec}
A.~Grover and J.~Leskovec.
\newblock node2vec: Scalable feature learning for networks.
\newblock In \emph{SIGKDD}, pages 855--864, 2016.

\bibitem[Henaff et~al.(2015)Henaff, Bruna, and LeCun]{henaff2015deep}
M.~Henaff, J.~Bruna, and Y.~LeCun.
\newblock Deep convolutional networks on graph-structured data.
\newblock \emph{arXiv:1506.05163}, 2015.

\bibitem[Holme and Saram{\"a}ki(2012)]{holme2012temporal}
P.~Holme and J.~Saram{\"a}ki.
\newblock Temporal networks.
\newblock \emph{Physics Reports}, 2012.

\bibitem[Kong et~al.(2012)Kong, Yu, Ding, and Wild]{kong2012meta}
X.~Kong, P.~S. Yu, Y.~Ding, and D.~J. Wild.
\newblock Meta path-based collective classification in heterogeneous
  information networks.
\newblock In \emph{CIKM}, pages 1567--1571. ACM, 2012.

\bibitem[Lee et~al.(2017)Lee, Rossi, and Kong]{lee17-Deep-Graph-Attention}
J.~B. Lee, R.~Rossi, and X.~Kong.
\newblock Deep graph attention model.
\newblock In \emph{arXiv:1709.06075}, pages 1--8, 2017.

\bibitem[Neville et~al.(2003)Neville, Jensen, Friedland, and
  Hay]{neville:kdd03}
J.~Neville, D.~Jensen, L.~Friedland, and M.~Hay.
\newblock Learning relational probability trees.
\newblock In \emph{SIGKDD}, pages 625--630, 2003.

\bibitem[Oyama et~al.(2011)Oyama, Hayashi, and Kashima]{oyama2011cross}
S.~Oyama, K.~Hayashi, and H.~Kashima.
\newblock Cross-temporal link prediction.
\newblock In \emph{ICDM}, pages 1188--1193. IEEE, 2011.

\bibitem[Park et~al.(2017)Park, Moore, and Neville]{parkdeep}
H.~Park, J.~Moore, and J.~Neville.
\newblock Deep dynamic relational classifiers: Exploiting dynamic neighborhoods
  in complex networks.
\newblock pages 1--7, 2017.

\bibitem[Perozzi et~al.(2014)Perozzi, Al-Rfou, and Skiena]{deepwalk}
B.~Perozzi, R.~Al-Rfou, and S.~Skiena.
\newblock Deepwalk: Online learning of social representations.
\newblock In \emph{SIGKDD}, pages 701--710, 2014.

\bibitem[Rossi and Ahmed(2015)]{roles2015-tkde}
R.~A. Rossi and N.~K. Ahmed.
\newblock Role discovery in networks.
\newblock \emph{Transactions on Knowledge and Data Engineering}, 27\penalty0
  (4):\penalty0 1112--1131, April 2015.
\newblock ISSN 1041-4347.
\newblock \doi{10.1109/TKDE.2014.2349913}.

\bibitem[Rossi et~al.(2012)Rossi, McDowell, Aha, and Neville]{rossi12jair}
R.~A. Rossi, L.~K. McDowell, D.~W. Aha, and J.~Neville.
\newblock Transforming graph data for statistical relational learning.
\newblock \emph{Journal of Artificial Intelligence Research}, 45\penalty0
  (1):\penalty0 363--441, 2012.

\bibitem[Rossi et~al.(2017)Rossi, Zhou, and Ahmed]{deepGL}
R.~A. Rossi, R.~Zhou, and N.~K. Ahmed.
\newblock Deep feature learning for graphs.
\newblock In \emph{arXiv:1704.08829}, 2017.

\bibitem[Singh et~al.(2007)Singh, Xu, and Berger]{singh2007-matching-topology}
R.~Singh, J.~Xu, and B.~Berger.
\newblock Pairwise global alignment of protein interaction networks by matching
  neighborhood topology.
\newblock In \emph{Proceedings of the 11th Annual International Conference on
  Research in Computational Molecular Biology}, volume 4453 of \emph{Lecture
  Notes in Computer Science}, pages 16--31. Springer, 2007.

\bibitem[Sun and Han(2013)]{sun2013mining}
Y.~Sun and J.~Han.
\newblock Mining heterogeneous information networks: a structural analysis
  approach.
\newblock \emph{ACM SIGKDD Explorations Newsletter}, 14\penalty0 (2):\penalty0
  20--28, 2013.

\bibitem[Sun et~al.(2012)Sun, Han, Yan, and Yu]{sun2012mining}
Y.~Sun, J.~Han, X.~Yan, and P.~S. Yu.
\newblock Mining knowledge from interconnected data: a heterogeneous
  information network analysis approach.
\newblock \emph{Proceedings of the VLDB Endowment}, 5\penalty0 (12):\penalty0
  2022--2023, 2012.

\bibitem[Tang et~al.(2015)Tang, Qu, Wang, Zhang, Yan, and Mei]{line}
J.~Tang, M.~Qu, M.~Wang, M.~Zhang, J.~Yan, and Q.~Mei.
\newblock {LINE: Large-scale Information Network Embedding}.
\newblock In \emph{WWW}, pages 1067--1077, 2015.

\bibitem[Viswanath et~al.(2009)Viswanath, Mislove, Cha, and
  Gummadi]{viswanath2009evolution}
B.~Viswanath, A.~Mislove, M.~Cha, and K.~P. Gummadi.
\newblock {On the evolution of user interaction in Facebook}.
\newblock In \emph{OSN}, pages 37--42, 2009.

\bibitem[Zager and Verghese(2008)]{zager2008graph}
L.~A. Zager and G.~C. Verghese.
\newblock Graph similarity scoring and matching.
\newblock \emph{Applied mathematics letters}, 21\penalty0 (1):\penalty0 86--94,
  2008.

\bibitem[Zhu et~al.(2003)Zhu, Ghahramani, and Lafferty]{zhu2003semi}
X.~Zhu, Z.~Ghahramani, and J.~D. Lafferty.
\newblock Semi-supervised learning using gaussian fields and harmonic
  functions.
\newblock In \emph{ICML}, pages 912--919, 2003.

\end{thebibliography}
\bibliographystyle{abbrvnat}

\end{document}